\documentclass[twocolumn,11pt]{article}

\usepackage[letterpaper,left=2cm,right=2cm,bottom=2cm,top=2cm]{geometry}

\usepackage{url}
\usepackage{cite}
\usepackage{amsmath,amssymb,amsfonts}
\usepackage{algorithmicx}
\usepackage{algorithm}
\usepackage{algpseudocode}
\usepackage{graphicx}
\usepackage{booktabs}
\usepackage{float}
\usepackage{multicol}
\usepackage{enumitem}

\usepackage{textcomp}
\usepackage{multirow}
\usepackage{xcolor}
\usepackage{adjustbox}
\usepackage[english]{babel}

\usepackage[utf8]{inputenc}
\usepackage[T1]{fontenc}
\usepackage{adjustbox}
\newcommand{\comment}[1]{}
\def\BibTeX{{\rm B\kern-.05em{\sc i\kern-.025em b}\kern-.08em
    T\kern-.1667em\lower.7ex\hbox{E}\kern-.125emX}}

\begin{document}
%\mainmatter              % start of a contribution
\title{\textbf{BERTaú: Itaú BERT for digital customer service}}

\author{   \texttt{Paulo Finardi} \and \texttt{José Dié Viegas} \and \texttt{Gustavo T. Ferreira} \and \texttt{Alex F. Mansano} \and \texttt{Vinicius F. Caridá} \\
\texttt{MaLS Data Science Team - Digital Customer Service, Itaú Unibanco, São Paulo, Brazil}}

\date{\texttt{email: \{paulo.finardi, jose.barros-viegas, gustavo.tino-ferreira, alex.mansano, vinicius.carida\}@itau-unibanco.com.br} \vspace{0.4cm} }

\maketitle

\begin{abstract}
In the last few years, three major topics received increased interest: deep learning, NLP and conversational agents. Bringing these three topics together to create an amazing digital customer experience and indeed deploy in production and solve real-world problems is something innovative and disruptive. We introduce a new Portuguese financial domain language representation model called BERTaú. BERTaú is an uncased BERT-base trained from scratch with data from the Itaú virtual assistant chatbot solution. The novelty of this contribution lies in that BERTaú pretrained language model requires less data, reaches state-of-the-art performance in three NLP tasks, and generates a smaller and lighter model that makes the deployment feasible. We developed three tasks to validate our model: information retrieval with Frequently Asked Questions (FAQ) from Itaú bank, sentiment analysis from our virtual assistant data, and a NER solution. All proposed tasks are real-world solutions in production on our environment and the usage of a specialist model proved to be effective when compared to \texttt{Google BERT multilingual} and the  Facebook’s \texttt{DPRQuestionEncoder}, available at Hugging Face. \texttt{BERTaú} improves the performance in $22\%$ of FAQ Retrieval MRR metric, $2.1\%$ in Sentiment Analysis F$_1$ score, $4.4\%$ in NER F$_1$ score. It can also represent the same sequence in up to $66\%$ fewer tokens when compared to "shelf models".
\end{abstract}

\section{Introduction}
In recent years, creating and managing digital customer experiences has appeared as to be a key area for many companies on “leveraging digital advancement for the growth of organizations and achieving sustained commercial success” \cite{Bones2017}. The idea of interacting with computers through natural language dates back to the 1960s, but recent technological advances have led to a renewed interest in conversational agents such as chatbots and digital assistants. In the customer service context, conversational agents promise to create a fast, convenient, and cost-effective channel for communicating with customers \cite{Gnewuch2017}.

In the area of digital customer service, there is a growing demand for assertiveness and specialization, since a more assertive service allows the customer to be served more quickly, thus increasing the efficiency of the entire system. In this context, three main standard NLP (natural language processing) tasks stand out, as follows:
1- Named Recognition Entity (NER): recognizing the entities during the conversation is fundamental to understanding customer's demands.
2- Information Retrieval (IR) with Frequently Asked Questions (FAQ): once the demand is understood, it is necessary to present the most relevant information.
3- Sentiment Analysis (SA): managing customer sentiment/satisfaction has become crucial\cite{Mousavi2020}.

Machine learning has received increased interest both as an academic research field and as a solution for real-world business problems. In benchmarks of NLP tasks, such as squad\cite{rajpurkar2016squad} and glue\cite{wang2019glue}, machine learning models can achieve better performance than humans. The BERT\cite{bert} algorithm and variants are considered state-of-the-art at solving NLP tasks. However, the deployment of machine learning models in production systems can present several issues and concerns \cite{paleyes2020challenges}.

Given the context described so far,
the goal of this paper is to test these hypothesized advantages of using and fine-tuning pretrained language models for a Brazilian Portuguese financial domain. To address these problems, we developed \texttt{BERTaú}, a BERT base uncased pretrained with data from AVI (Itaú Virtual Assistant).

Given that the raw text is at the core of digital service data, it is necessary to use increasingly robust models to achieve the best possible digital service.

Our choice for using BERT \cite{bert} is based on the idea that it is an encoder model that is powerful and widely established in the Natural Language Processing (NLP) field and programming libraries, such as Hugging Face Transformers \cite{huggingfaces}.Also, Open NeuralNetwork Exchange (ONNX) \cite{onnx} has presented good deployment solutions and tools for reducing inference time, which allows \texttt{BERTaú} to be deployed.

The key point of this research is the data. For better results, it is mandatory to have a big and high-quality dataset - a large number of tokens with good semantic text. From the fact that the AVI has an average of 2 million monthly digital visits, few months of data would be necessary to enable it to train the model from scratch. However, we used a larger data window to capture semantic text before and after the pandemic $-$ approximately 18 months of data. Further details about the data and dataset is described in section \ref{sec:dataset}.

Once the model has been trained, we validate the model in three main tasks in our environment: Information Retrieval (IR) with Frequently Asked Questions (FAQ), Sentiment Analysis (SA) on phrases about our service, with three classes: positive, neutral and negative, and Named Recognition Entity (NER), which recognizes entities during the conversation between the customer and AVI.

The article is organized as follows: in section~\ref{sec:dataset} we describe the process of the training from scratch; in section~\ref{sec:baselines} we detail the baselines and related work; in section \ref{sec:experiments} we describe the setup for our experiments; section~\ref{sec:experiments} presents the results; and finally, in section~\ref{sec:conclusion} we describe our evaluation results and conclusions.

\section{BERTaú From Scratch}
\label{sec:dataset}
In this work, we chose to train BERT uncased, since the vast majority of the dataset that constitutes AVI conversations are in this format. We have $14,5$GB of data ($22,500,000$ words). Each line in the data is an AVI session, that is: A complete iteration between customer and chatbot. For obvious reasons, we swapped all numbers in the dataset with random numbers, in order to avoid any possibility of sensitive information, document number, or monetary values being exposed. Regarding the training configuration, we follow the guidelines of the BERT article in a straight line, with a maximum sequence length of $512$ tokens and batch size of $256$ sequences. The model was trained for $1,000,000$ steps with a learning rate of $5$e-$5$ and a warm up of $20,000$ steps. The accuracy and loss at the end of training is shown in Table \ref{tab:results_training}.

\subsection{Vocabulary}
The dataset for the vocabulary creation was also extracted from the AVI, but with a different period \footnote{without interception to the data used in BERT training}. This dataset has $2$GB and $34100$ tokens. We create our vocabulary using the Hugging Face tokenizer \texttt{BertWordPieceTokenizer}, performing \texttt{NFKC} text normalization, and setting \texttt{strip\_accents} to \texttt{false}. 

\begin{table}[htb!]
\centering\centering\resizebox{0.5\textwidth}{!}{
\begin{tabular}{lcc}
\toprule
% \toprule
\textbf{Eval training on step 1,000,000} & \textbf{Accuracy} & \textbf{Loss} \\ 
\midrule
\texttt{Masked Language Model}          & 0.785     & 0.891 \\
\texttt{Next Sentence Prediction}       & 0.863     & 0.311 \\
\bottomrule
\end{tabular}
}
\caption{Evaluation training accuracy and loss of Masked Language Model (MLM) and Next Sentence Prediction (NSP) tasks.}
\label{tab:results_training}
\vspace{-4mm}
\end{table}

\section{Baselines and Related Work}
\label{sec:baselines}
Although the BERT model is open source and offers pretrained weights in English language and has a multilingual version available (\texttt{mBERT}), many studies have been developed to create specialist BERTs. A natural advantage of having a model pretrained in a specific vocabulary domain is that it can represent sequences using fewer tokens and performs the training stage with more computational efficiency. Table \ref{tab:med_token} directly compares the sequence lengths between our model and the baselines in NER and Information Retrieval tasks. Besides short token representations, many specialized models have reached the state of the art (sota) when trained from scratch, the \texttt{BERTimbau}\footnote{BERTimbau results at: \url{https://github.com/neuralmind-ai/portuguese-bert}} \cite{bertimbau} is sota on NER task at public Portuguese MiniHAREM dataset. In the financial domain, the \texttt{FinBERT} \cite{finbert} achieves sota on financial sentiment analysis.

\begin{table}[htb!]
\centering\centering\resizebox{0.5\textwidth}{!}{
\begin{tabular}{lcc}
\toprule
\textbf{Model} & \textbf{FAQ$_{\text{dataset}}$} & \textbf{NER$_{\text{dataset}}$}  \\ 
\midrule
\texttt{DPRQuestionEncoder}                    & 130 tokens    & - \\
\texttt{mBERT uncased}                         & 97 tokens     & 13 tokens \\
\texttt{BERTimbau - Portuguese BERT}           & 90 tokens     & 12 tokens \\
\texttt{BERTaú}                                & 78 tokens     & 10 tokens \\
\bottomrule
\end{tabular}
}
\caption{Comparison of token lengths between models, where the models in FAQ dataset vary between [$15\%$, $66\%$] of sequences sizes greater than \texttt{BERTaú} and in the NER dataset between [$20\%$, $30\%$].}
\label{tab:med_token}
\end{table}

In the NER task we compared our model with \texttt{BERTimbau} base and \texttt{mBERT} uncased. In Information Retrieval task we conducted experiments where we compared our model with \texttt{BM25+} algorithm \cite{bm25plus}, Sentence Transformer (SBERT) -  \texttt{distiluse-base-multilingual} \cite{sentence_transformers},  Dense Passage Retrieval (\texttt{DPR}) QuestionEncoder \cite{dpr} and the \texttt{mBERT} model \cite{bert}.

\subsubsection{BM25+}
The \texttt{BM25} is a probabilistic model meant to estimate the relevance of a document based on the idea that the query terms have different distributions in relevant and non-relevant documents. \texttt{BM25+} has a better way of scoring long documents with the addition of a parameter $\delta$ that has a default value of $1$. The formula is
\begin{equation}
\begin{split}
    score(D,Q) & = \sum_{i=1}^{n} IDF(Q_i) \bigg[  \\ 
    & \dfrac{TF(Q_i, D)\cdot (k_1 + 1)}{TF(Q_i, D)+ k_1 \cdot \left(1-b+b\cdot \frac{|D|}{\text{avgdl}} \right)} + \delta \bigg], 
\end{split}
\end{equation}
where $Q_i$ is term frequency in the document $D$, $|D|$ is the length of the document $D$ in words, and avgdl is the average document length in the text collection. $k_1$ and $b$ are free parameters. More details about \texttt{BM25+} are shown in \cite{bm25plus} section 3.3.

\subsubsection{SBERT - distiluse-base-multilingual}
We use the \texttt{distiluse-base-multilingual}\footnote{Model available at: \url{https://www.sbert.net/examples/training/multilingual}}
model proposed by \cite{sentence_transformers} and measure the similarity between the question and answer by cosine distance. The distiluse model supports $50+$ languages including Portuguese.

\subsubsection{Dense Passage Retrieval (DPR) QuestionEncoder}
The \texttt{DPR} \cite{dpr} has the same structure as BERT and tackles the problem with the objective of getting better representations of dense embeddings. This model was trained in pairs of questions and answers. The authors also argue that the best way to obtain better representations of dense embeddings for Information Retrieval task is by maximizing the inner products of the question and vectors of relevant answers in a batch.  

\subsubsection{mBERT uncased}
The best-known \texttt{mBERT} model is the multilingual BERT \cite{bert} and was used as a baseline for Information Retrieval, NER and Sentiment Analysis tasks, this model was trained on the XNLI: Cross-Lingual NLI corpus \cite{xnli} with $102$ languages.

\section{Experiments}
\label{sec:experiments}
Another feature of BERT is that it is a versatile model. The structure of the Transformer encoder allows the model to perform different tasks with few adaptations. Given its versatility, we perform three different experiments with \texttt{BERTaú}:
\begin{enumerate}[noitemsep]
    \item FAQ retrieval 
    \item Sentiment Analisys (SA)
    \item Named Entity Recognition (NER)
\end{enumerate}

\noindent For all experiments, we used Nvidia GPU's with half-precision \texttt{FP16}, which allows for less memory use during the training phase. Using \texttt{FP16} with the same batch size, we were able to perform experiments twice as fast when compared to the \texttt{FP32} full precision. A minor drawback of \texttt{FP16} is its limited range: large numbers can explode and small numbers are truncated to zero. To avoid this behavior it is common to scale the objective function by a large number in the gradients. If this number is so large that it generates an overflow, the update is rejected and a new scale factor is obtained automatically with PyTorch Automatic Mixed Precision.

\subsection{FAQ Retrieval}
\begin{figure}[htb!]
\centering
\includegraphics[width=1\linewidth]{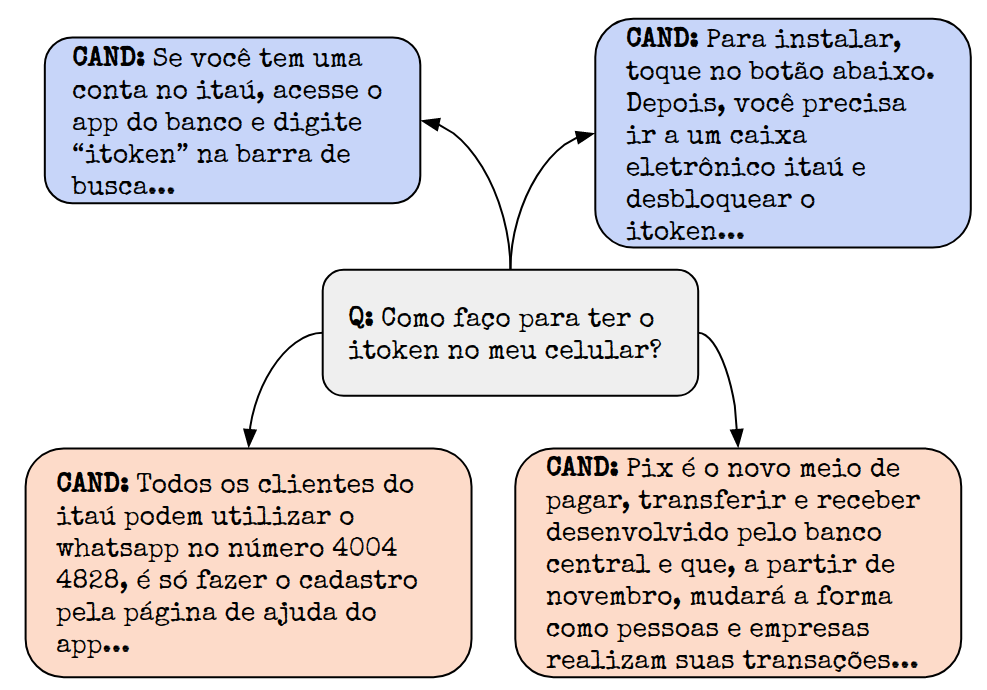}
\caption{FAQ retrieval dataset sample, where oranges CANDs are wrong and blue CANDs the correct answers for Q.}
\label{faq}
\end{figure}

The task of FAQ retrieval can be described as follows: \textit{Given a question and a set of candidate answers we want to identify which are the correct answers / true labels}. This is a classic problem in the Information Retrieval field and many solutions have been developed in the last $3$ years. A complete survey with this kind of solutions can be found here \cite{irsurvey}. Regarding metrics, we use the Reciprocal Rank (RR). The RR of a single query is given by $\frac{1}{rank_K}$, where rank is a descending ordered list of size $K$: where answers with greater probabilities appear at the beginning of the list. The  Reciprocal Rank is meant to answer: "\textit{Where is the first relevant answer on the list?}", for example: if the relevant answer gets the seventh place, the RR for this query is $1/7$ and if no answer is found, the RR is $0$. To calculate the RR for multiple queries we use the \texttt{MRR} (Mean Reciprocal Rank) metric, which is the RR average and given by the formula:
\begin{equation}
MRR = \dfrac{1}{Q}\sum_{j=1}^{Q}\dfrac{1}{rank_j}
\end{equation}
We also measure the experiments with Average Precision \texttt{AP@1} in the first position,  which determines whether the first answer on the list is correct or not. The dataset was obtained from the FAQ Itaú bank\footnote{A sample set of FAQs used to build the dataset can be found here: \url{https://www.itau.com.br/contas/conta-corrente/}} . Our data has $1427$ questions with $2118$ answers and some questions have more than one correct answer, i.e. the question $\mathbf{Q_1}$ can be an answer of $\{\mathbf{A_1, A_2}, \dots, \mathbf{A_n}\}~ n \leq 5$ and the average sequence length $\mathbf{QA}$ pair is $192$ tokens. The Figure \ref{faq} shows one FAQ data sample.

Our approach for this problem is pretty simple: we tackle the problem as a binary classification problem, where the samples in the dataset are triples $\{( \mathbf{Q_1,C_1, A_1}), (\mathbf{Q_2,C_2, A_2}), \dots, (\mathbf{Q_n,C_n, A_n}) \}$, where $\mathbf{C_k}$ with $k=1,2,\dots, n$ is a list of candidates answers chosen randomly. For each triple $(\mathbf{Q_k,C_k, A_k})$ we create $M$ training samples, where $M \in \{15, 30, 45 \}$ and this method follows the same idea introduced in  \cite{reviewRanking}. After training the binary classifier, we rank the answers according to the Algorithm \ref{alg:faq}. For the same dataset and ranking proposal we use two approaches: The pointwise and pairwise \cite{liu2011learning}. 

\begin{algorithm}
\caption{Ranking FAQ}\label{alg:faq}
\begin{algorithmic}[1]
\State \textbf{Input:} logits from output model
\State \textbf{Output:} dict q$_{\text{id}} \rightarrow$ list(predict\_rank)
\For{cand in cands}
    \State logits = output-model
    \State pred\_values = softmax(logits) 
    \State pred\_index = argsort(pred\_values)
    \State predict\_rank = doc$_{\text{ids}}$[pred\_index]
\EndFor
\end{algorithmic}
\end{algorithm}

\subsubsection{Pointwise - Label Smoothing}
Our pointwise method uses label smoothing \cite{label_smoothing} as an objective function to regularize the neural network by penalizing confident output distribution. The label smoothing mitigates the symptom of overfitting by penalizing output distributions with low entropy. Low entropy occurs when the network places all probability on a single class during the training phase \cite{label_smoothing_1}. We set up (without any grid search) the confidence penalty to $0.1$. 

Alternatively, we have also experimented with Cross Entropy loss (when setting the confidence penalty to 0). This experiment had slightly lower results in \texttt{MRR@10} and \texttt{AP@1}.

\subsubsection{Pairwise}
\begin{figure}
\centering
\includegraphics[width=1\linewidth]{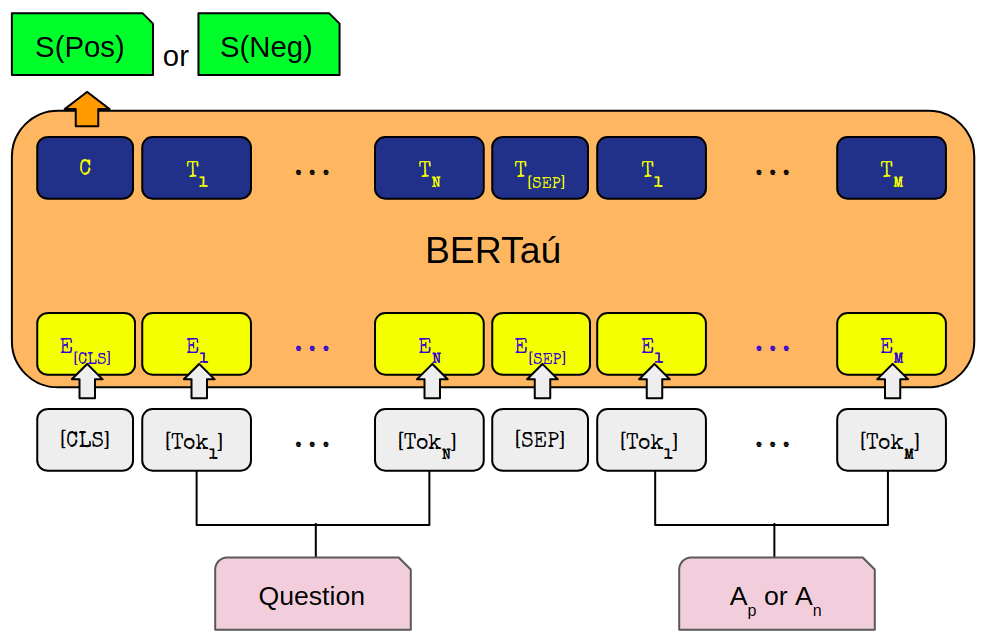}
\caption{Pairwise model struct, adapted from \cite{bertsel}.}
\label{fig:bertsel}
\end{figure}
The pairwise approach follows the same one adopted in \cite{bertsel} and shown in Figure \ref{fig:bertsel} which takes a couple of candidate answers and learns the most relevant answer for the question. Explicitly, he have  a triple $(\mathbf{Q, A_p, A_n})$ where $\mathbf{Q}$ is the question, $\mathbf{A_p}$ and  $\mathbf{A_n}$ are the positive and negative answer respectively. This triple is broken into two pairs $(\mathbf{Q, A_p})$ and $(\mathbf{Q, A_n})$ and each pair is sent individually to a \texttt{BERTaú}. For the loss function, we used only the hinge loss pairwise function and not the combined cross entropy loss function with hinge loss function proposed in \cite{bertsel} because the two strategies obtained similar results and we explicitly choose the simplest loss. The loss used in the pairwise model is:
\begin{equation}
\max\{0, M  -  \hat{y}_\theta(\mathbf{Q, A_p}) + \hat{y}_\theta(\mathbf{Q, A_n})\}    
\end{equation}
where $\hat{y}_\theta(\mathbf{Q, A_p})$ and  $\hat{y}_\theta(\mathbf{Q, A_n})$ denote the predicted scores of positive and negative answer, whereas $M$ is the margin parameter, in our case fixed in $M=0.2$. 

\begin{figure}[htb!]
\centering
\includegraphics[width=1\linewidth]{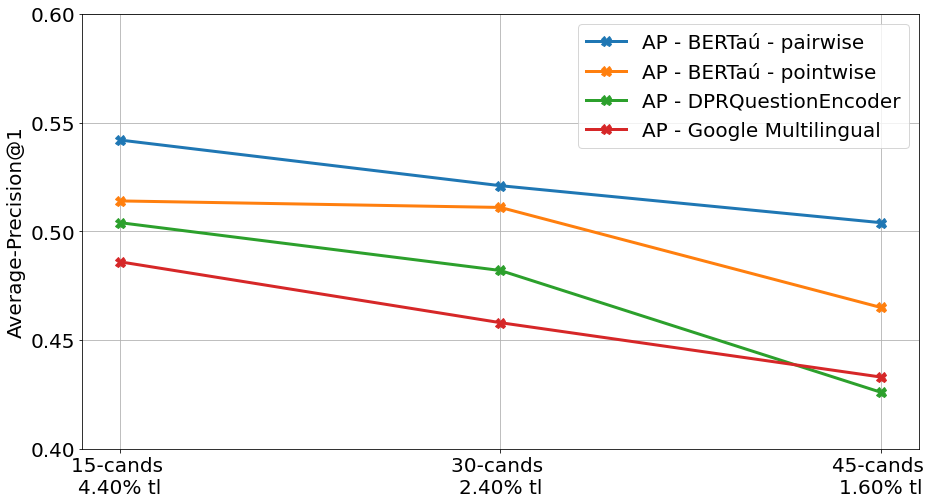} 
\caption{Comparison between models, varying the imbalance of the FAQ dataset. x-axis legend means the number of possible candidates for one sample and xx\%-tl is the percentage of unbalance in the dataset where tl = true-label.}
\label{fig:av1}
\end{figure}

For the FAQ retrieval experiment we used AdamW optimizer, learning rate of  $5$e-$5$, linear scheduler with warmup of $2\%$ of total steps\footnote{total steps = \texttt{(len(train\_loader) * \# epochs)}} for $1$ epoch.\footnote{We have tested for $2$ and 3 epochs, obtaining similar results but with more overfit symptoms}  We also varied the number of cands in the dataset build, by increasing the number of cands for the same question, the imbalance of the dataset increases, we testing three levels of cands \{15,30,45\}, the Figure \ref{fig:av1} shows the performance of the pointwise and pairwise models. Table \ref{tab:results_faq} shows the main FAQ retrieval results. Note that the pairwise model gets the best results, however this model takes twice as long in training when compared with pointwise.

\begin{table}[htb!]
\centering\centering\resizebox{0.5\textwidth}{!}{
\begin{tabular}{lcc}
\toprule
\textbf{Model} & \textbf{MRR@10} & \textbf{AP@1}  \\ 
\midrule
 
\texttt{BM25+}                                 & 0.345     & 0.246 \\
\texttt{distiluse-base-multilingual}           & 0.417     & 0.285 \\
\texttt{mBERT uncased}                         & 0.458     & 0.505 \\
\texttt{DPRQuestionEncoder}                    & 0.526     & 0.482 \\
\texttt{BERTaú pointwise label smoothing}      & 0.544     & 0.511 \\
\texttt{BERTaú pairwise}                       & \textbf{0.552}     & \textbf{0.521} \\
\bottomrule
\end{tabular}
    }
\caption{Evaluation result on IR FAQ dataset for each sample with 30 candidates  unbalanced dataset: 97.60/2.40 true labels.}
\label{tab:results_faq}
\end{table}

\subsection{Sentiment Analysis}

% \footnote{BERTimbau results at: \url{https://github.com/neuralmind-ai/portuguese-bert}}
The study of Sentiment Analysis, brings valuable feedbacks for any kind of problems during the digital customer service . When we identify the feeling of a sentence, We can act proactively and guide customer service more efficiently.

In this sense, it is essential to promptly identify customers who demonstrate dissatisfaction, so that they can count on the proper assistance for their respective issues. The institution must also understand topics that have good acceptance so that it can ensure suitable maintenance and continuous improvement. Finally, its technicians must also fix processes and functionalities that are often related to negative sentiments. Thus, offline models of sentiment analysis were developed with the help of machine learning techniques.

%To act on this agenda in the context of Itaú-Unibanco's virtual assistant (AVI), offline models of sentiment analysis were developed with the help of machine learning techniques. 

%Secondary objectives of the model are related to the understanding of intentions that have good acceptance for maintenance and continuous improvement, as well as fixing processes and functionalities that are often related to negative feelings.

%For empirical reasons, instead of a binary model with positive and negative classes for sentiments, the classification metrics proved to be more robust for multiclass models with three classes of sentiment: positive, negative and neutral.

Given the considerable amount of interactions that may not express any sentiment, the classification metrics proved to be more robust for multiclass models with three classes of sentiment: positive, negative and neutral:

%Initially, in order to obtain the labeled data, the team of data scientists were assisted by the data quality team.

%Since human labeling can be subjective, some criteria have been defined that characterize these 3 classes accurately:

\begin{enumerate}
    \item Positive class - Praise and thanks.
    \item Neutral Class - Doubts, requests and tracking of status of preview demands.
    \item Negative Class - Cursing, complaints and indication of complaints in consumer protection agencies, reports of systemic errors, indication of improper service, disagreement with products, services and fees. 
\end{enumerate}

These classes follow the pattern of classical literature for the problem of sentiment analysis \cite{sent-ref}. Also, aiming to expand the labeled dataset reliably, databases from other sources and service channels were used, such as the bank's official social networks, examples of previously labeled interactions from human chat, and also telephone transcriptions.

\subsubsection{Pre-processing}

During the development of these models, a relatively traditional pre-processing text approach was used.

\begin{enumerate}
    \item Removing accents, punctuation and special characters.
    \item Conversion to lowercase. 
    \item Removing numbers. 
    \item Removing stopwords from a validated dictionary in the business context.
\end{enumerate}

For comparison purposes, it is also worth noting that the models were trained by alternating stemming.

%Finally, for the vector representation of the documents, in the context of this paper, the individual interactions of the clients in the virtual assistant (AVI), trigrams were used.

For a representation of the term-document matrix, we decided to use n-grams, more precisely unigrams, bigrams and trigrams, where, each term $ t_j $ (n-gram) of each document $ d_i $, the product $ tfidf (t_j, d_i) = TF (t_j, d_i) IDF (t_j) $ was computed. The $ TF (t_j, d_i) $ can be interpreted as the frequency of occurrence of the n-gram $ t_j $ in a $ d_i $ document, and $ IDF (t_j) $ as a measure of occurrence of the term $ t_j $ in the corpus as a whole.

Thus, from all n-grams $t_j $, $ i = (1, \hdots, M)$ and all documents $ d_i $, $ i = (1, \hdots, N)$, we compute the products $ TF ( t_j, d_i). IDF (t_j)$ and form the matrix $T$.

To address issues related to the sparsity and high dimensionality of the corpus, we performed a \textit{Singular Value Decomposition} on the matrix $T$ (or SVD, for short).

Therefore, $T$ can be represented as:

\begin{eqnarray}
T' = U  \Sigma V^{t},
\end{eqnarray}  
where $U$ contains the eigenvectors of the correlations of the terms $t$, $V$ contains the eigenvectors of the correlations of the documents $d$ and $\Sigma$ contains the singular values of the decomposition.

To find the projection of the vector representation of a document $\textbf{d}_j$  in the new space, we can write: \\
\begin{eqnarray}
\hat{\textbf{d}_j} = \Sigma^{-1} U^{t} \textbf{d}_j
\end{eqnarray}

We applied a grid-search to find the best hyperparameters and ideal number of components used in SVD.

%he construction of the model, when the training pipeline was started, in addition to a grid-search to find the optimal hyperparameters of each model, a grid-search was also performed for the number of principal components.

In the best model obtained, the initial 64,686 features were reduced via SVD decomposition to 650 components. The stage of decomposition into singular values was the most computationally expensive, even though it was performed using parallelization.

\subsubsection{Semi-Supervised Learning}
As it is an intrinsically human activity, the labeling phase by the data quality team is, in general, a moment in the process of creating models that naturally demands a longer execution time. For this step, each text example (document) selected to be labeled was examined by 2 different analysts, so that labeling errors due to individual bias were mitigated. 

Only those documents that had been labeled equally by different analysts were selected to compose the training set. This way of composing the training set, although guaranteeing higher quality, also reduced drastically the example labeling rate.

%Thus, at a certain point, in order to test some hypotheses and new attendance pipelines, predictions from the sentiment analysis model were required and it was necessary to expand the training dataset more quickly.

In order to minimize this drawback and improve the labeling time, it was decided to use semi-supervised learning algorithms, in particular, the Co-training method \cite{blum98}.

The main idea is based on the cooperation of two supervised learning algorithms. Be matrix $T$ with $N$ documents and $M$ features. %In addition, we will represent the negative, positive and neutral class with $-1$, $1$ and $0$, respectively. Here, the class labels represented by $?$ are unknown.

\comment{
\begin{blockarray}{cccccccc}
t_1 & t_2 & \hdots & t_j & \hdots & t_M & Y \\
\begin{block}{(ccccccc)c}
  a_{11} & a_{12} & \hdots & a_{1j} & \hdots & a_{1M} & 0 & d_1 \\
  a_{21} & a_{22} & \hdots & a_{2j} & \hdots & a_{2M} &-1 & d_2 \\
  a_{31} & a_{32} & \hdots & a_{3j} & \hdots & a_{3M}&1 & d_3 \\
   a_{41} & a_{42} & \hdots & a_{4j} & \hdots & a_{4M} & ? & d_4 \\
   a_{51} & a_{52} & \hdots & a_{5j} & \hdots & a_{5M} & ? & d_5 \\
  \vdots & \vdots & \hdots & \vdots & \hdots& \vdots  & \vdots & \vdots \\
  a_{N-1,1} & a_{N-1,2} & \hdots &
 a_{N-1,j} & \hdots & a_{N-1,M}  & y_{N-1} & d_{N-1} \\
  a_{N,1} & a_{N,2} & \hdots & a_{N,j} & \hdots & a_{N,M}  & y_{N} & d_{N} \\
\end{block}
\end{blockarray}
}

For the implementation of the Co-training method, the matrix $T$ was divided into two other matrices $T_1$ and $T_2$, as follows:
\begin{eqnarray}
T = T_1 \cup T_2\\
\emptyset = T_1 \cap T_2
\end{eqnarray}

If we think of each term $\vec{t_k}$ as a column vector of $T$, we can define $T_1$ and $T_2$ as:

\begin{eqnarray}
T_1 = (\vec{t_1}, \quad \vec{t_2}, \hdots, \quad \vec{t_j})\\
T_2 = (\vec{t_{j+1}}, \quad \vec{t_{j+2}}, \hdots,\quad \vec{t_M}),
\end{eqnarray}

\comment{
So, we can infer that $T_1$ is given by:

\begin{blockarray}{cccccc}
t_1 & t_2 & \hdots & t_j & Y \\
\begin{block}{(ccccc)c}
  a_{11} & a_{12} & \hdots & a_{1j}  & 0 & d_1 \\
  a_{21} & a_{22} & \hdots & a_{2j}  &-1 & d_2 \\
  a_{31} & a_{32} & \hdots & a_{3j}  &1 & d_3 \\
   a_{41} & a_{42} & \hdots & a_{4j} & ? & d_4 \\
   a_{51} & a_{52} & \hdots & a_{5j} & ? & d_5 \\
  \vdots & \vdots & \hdots & \vdots  & \vdots & \vdots \\
  a_{N-1,1} & a_{N-1,2} & \hdots &
 a_{N-1,j}   & y_{N-1} & d_{N-1} \\
  a_{N,1} & a_{N,2} & \hdots & a_{N,j} & y_{N} & d_{N} \\
\end{block}
\end{blockarray}
, and $T_2$ by:\\
 
\begin{blockarray}{cccccc}
t_{m+1} & t_{m+2} & \hdots & t_M & Y \\
\begin{block}{(ccccc)c}
  a_{1, m+1} & a_{1,m+2} & \hdots & a_{1M}  & 0 & d_1 \\
  a_{2, m+1} & a_{2,m+2} & \hdots & a_{2M}  &-1 & d_2 \\
  a_{3,m+1} & a_{3,m+2} & \hdots & a_{3M}  &1 & d_3 \\
   a_{4,m+1} & a_{4,m+2} & \hdots & a_{4M} & ? & d_4 \\
   a_{5,m+1} & a_{5,m+2} & \hdots & a_{5M} & ? & d_5 \\
  \vdots & \vdots & \hdots & \vdots  & \vdots & \vdots \\
  a_{N-1,m+1} & a_{N-1,m+2} & \hdots &
 a_{N-1,M}   & y_{N-1} & d_{N-1} \\
  a_{N,m+1} & a_{N,m+2} & \hdots & a_{N,M} & y_{N} & d_{N} \\
\end{block}
\end{blockarray}
}
so both $T_1$ and $T_2$ represent, somehow, the same documents, however with different sets of features.

For each of the sets, labeled examples were selected and the models $h_1$ and $h_2$ were trained. The techniques for training  $h_1$ and $h_2$ do not need to be exactly the same. However, for convenience of implementation, we chose the LightGBM \cite{Ke2017} method for both $h_1$ and $h_2$.
 
From $h_1$, unlabeled instances were classified into $T_1$ and the same process was repeated for unlabeled instances in $T_2$. At each step $k$ of the model, the sets $T_1$ and $T_2$ can be subdivided into ($L_{T_1}^k$, $U_ {T_1}^k$) and ($L_ {T_2}^k$, $U_ {T_2}^k$), respectively, where the set $L_ {T_i}^k$ contains the labeled samples  from $T_i$ of the $k_{th}$ interaction and ${U_ {T_i}}^k$ the unlabeled samples from $T_i$ in the $k_{th}$ interaction.
 
Given some objective criteria (see \cite{blum98}), with certain degree of confidence, once the combined predictions from $h_1^k$ and $h_2^k$ indicate that some unlabeled document $d_h$ belongs to a certain class, in the next interaction of the method, $d_h$, originally unlabeled, now comprises both $L_{T_1}^{k + 1}$ and $L_{T_2}^{k + 1}$.
 
This method is repeated until all instances are labeled or any stop criteria are met. %, according to the user who implemented the technique.
 
It is possible to show  \cite{blum98} that from $h_1$ and $h_2$  a third classifier can be formed, which works similarly to the Naive Bayes one.
 
The number of examples obtained from other service channels, and manually labeled was about $5k$ training instances. With the co-training method described above, it was possible to expand the dataset at least $5$ times.
 
%The human labeling added to the training instances obtained from other service channels included around $5k$ training instances.
% With the co-training method described above, it was possible to expand the dataset at least 5 times.

\subsubsection{Model}
Different types of models were tested and trained to classify sentiments for the AVI interactions expanded dataset.

The model that had the best F$_1$ score was a random forest with $450$ trees and a depth of $5$. The F$_1$ score for this configuration was $0.76$.

\subsubsection{Further considerations}
Other types of vector representations for AVI text interactions were also tested, in particular a $300$- component pretrained Portuguese skip-gram word2vec embedding that was extracted from a public repository. 

The results, however, were relatively worse than those obtained by means of  aforementioned method, a fact that seems to be related to the banking context present in the training dataset.

As previously emphasized, some criteria were used for labeling data under human supervision. Thus, when classification elements were found that were consistent with the negative class, it was prioritized, although - for example- the interaction could also bring a simple question (such as the status of a credit card request), praise or thanks. This orientation aimed to ensure that interactions with a negative sentiment were not labeled in other classes because they also could satisfy some of the rules contemplated by these other classes, a fact that implies a certain hierarchy between the classes.

It is worth noting that in order to achieve even more confidence in the classification of sentiments, some interactions are previously identified via an exact match and also by (high) similarity methods that use Bag-of-Words representations and compute both the distances, cosines and Levensthein. The main results of the SA task are shown in Table \ref{tab:results_sa}.

\begin{table}[htb!]
\centering\centering\resizebox{0.5\textwidth}{!}{
\begin{tabular}{lcc}
\toprule
% \toprule
\textbf{SA task evaluated on F$_1$ score} & \textbf{ SA$_{\texttt{trinary}}$} & \textbf{ SA$_{\texttt{binary}}$} \\ 
\midrule
\texttt{Random Forest + SVD + n-grams}     & 0.760          & -\\
\texttt{mBERT uncased}                      & 0.838          & 0.901\\
\texttt{BERTaú}                             & \textbf{0.850} & \textbf{0.920}\\
\bottomrule
\end{tabular}
}
\caption{Evaluation result on SA task. }
\label{tab:results_sa}
\end{table}

\subsection{Named Entity Recognition}

Named Entity Recognition (NER) is one of the most common and fundamental sequence labeling tasks. Given a text, the objective is to correctly identify and extract a set of general-purpose and domain-specific named entities on a token level. While previous works focus on combining \texttt{LSTM} (Long Short-Term Memory) \cite{hochreiter1997long} and \texttt{CRF} (Conditional Random Fields) \cite{lafferty01Crf} models, our approach is to attach a dense layer at the end of the 's model and train it to predict each token's entity class. 
% TINO: falta citação dos trabalgos com lstm e crf.

Our NER dataset consists of $18370$ manually annotated examples. Each label is, then,  further divided following the BILOU schema (\texttt{B - `beginning' I - `inside' L - `last' O - `outside' U - `unit'}) with $16$ different classes. The classes include specific banking products, services, functionalities, organizations, companies, places, and documents.

In our experiments, we tested three different ways of using \texttt{BERTaú's} weight vectors: concatenating the last $4$ layers, summing the last $4$ layers, and using the last layer only. The results were very similar, with the summing approach being slightly better. The best configuration was a dense layer trained with AdamW optimizer using a learning rate of $5\times 10^{-5}$ with linear scheduler, and a warm-up of $2\%$ of total steps for $5$ epochs. 
We measured the performance of the models using seqeval's \cite{seqeval} implementation of the $F_1$ score, due to some classes being severely unbalanced. The results are shown in Table \ref{tab:results_ner}.
% TINO: falta citação do adam, talvez?

% Given a text passage the objective of the NER task is to correctly identify and classify entities. Our NER dataset has $18370$ examples with the BILOU schema \texttt{B - 'beginning' I - 'inside' L - 'last' O - 'outside' U - 'unit'} with $16$ classes divided into: \{boleto, canal, categoria, dados, data, dinheiro, documento, funcionalidade, lugar, organizacao, outros, pessoa, produto, servico, status, tarifa\}. For the experiment we tested three points of \texttt{BERTaú} weights to perform the experiment: concatenation of the last $4$ layers, sum of the last $4$ layers and use the last layer. The results, The results were similar, slightly better for using the sum of the last $4$ layers. The final experiment configuration is using AdamW optimizer, learning rate of $5$e-$5$ with linear scheduler with warmup of $2\%$ of total steps for $5$ epochs. The evaluation of the experiment was measured  with seqeval library  \cite{seqeval} with on the $F_1$ score. The results are shown in Table \ref{tab:results_ner}.% 

\begin{table}[htb!]
\centering\centering\resizebox{0.47\textwidth}{!}{
\begin{tabular}{lc}
\toprule
% \toprule
\textbf{NER task evaluated with seqeval} & \textbf{F$_1$ score}  \\ 
\midrule
\texttt{mBERT uncased}                      & 0.840  \\
\texttt{BERTimbau - base}                   & 0.853  \\
\texttt{BERTaú}                             & \textbf{0.877} \\
\bottomrule
\end{tabular}
}
\caption{Evaluation result on NER task.}
\label{tab:results_ner}
\vspace{-4mm}
\end{table}

\subsection{Quantization}
In order to use our model as a feature extractor, i.e. to generate embedding representations of words and sentences from \texttt{BERTaú} we conducted our model to a simple quantization, although there are more sophisticated solutions with greater compression power such as the Open Neural Network Exchange (ONNX) \cite{onnx} we chose a more traditional and simple way, the PytTorch quantization in \texttt{INT8}, which converts the \texttt{FP32} tensors to \texttt{INT8}. A quantized model allows the 
storing tensors at lower bitwidths than floating point, which implies less time of inference. We compared the inference time and sizes on three devices: \texttt{GPU-V100}, \texttt{CPU} and \texttt{Quantized}. The results are in Table \ref{tab:quant}.
\begin{table}[htb!]
\centering\centering\resizebox{0.5\textwidth}{!}{
\begin{tabular}{llc}
\toprule
% \toprule
\textbf{Experiment} & \textbf{Device} & \textbf{Inference time} \\ 
\midrule
                        & \texttt{GPU - V100}   & 0.2129s \\
\textbf{FAQ Retrieval } & \texttt{Quantized}    & 7.2901s \\
                        & \texttt{CPU}          & 18.1826s \\
\midrule
                        & \texttt{GPU - V100}   & 0.01644s \\
\textbf{SA}             & \texttt{Quantized}    & 0.09357s \\
                        & \texttt{CPU}          & 0.15289s \\
\midrule
                        & \texttt{GPU - V100}  & 0.01944s \\
\textbf{NER}            & \texttt{Quantized}   & 0.03457s \\
                        & \texttt{CPU}         & 0.10064s \\
\bottomrule
\end{tabular}
}
\caption{GPU and CPU model size is $428.95$Mb, Quantized model is $183.76$Mb. Inference time in FAQ Retrieval experiment with \texttt{BERTaú} pointwise - label smoothing in one sample with $15$ cands and inference time of NER and SA experiments in one sample.}
\label{tab:quant}
\end{table}

\section{Conflicts of Interest}

Any opinions, findings, and conclusions expressed in this manuscript are those of the authors and do not necessarily reflect the views, official policies nor position of Itaú Unibanco.

\section{Conclusion}
\label{sec:conclusion}
The field of Deep Learning and NLP has developed rapidly and has achieved good results in several tasks in the academic universe. However, applying these solutions in the industry to solve real-world problems is still a challenge. In order to improve the efficiency of our digital customer service, it is necessary to seek better solutions and algorithms such as BERT, but also customized solutions that are feasible to be implemented in production ensuring good results. In addition, such results translate into actual improvements in the users' experience. 

In this work, we present BERTaú, a specialist pretrained BERT model in the AVI data and fine-tuned for three tasks: FAQ Retrieval, NER and Sentiment Analysis. when compared with the \texttt{mBERT} model, \texttt{BERTaú} improves the performance in $22\%$ of FAQ Retrieval \texttt{MRR@10}, $2.1\%$ in Sentiment Analysis F$_1$ score and $4.4\%$ in NER F$_1$ score. \texttt{BERTaú} can also represent the same sequence in up to $66\%$ fewer tokens when compared to "shelf models". This new approach, in addition to being applied to the three NLP tasks already carried out, we intend to apply our model as well as extracting resources from words and phrases. 

Due to responsibilities related to sensitive data, privacy and business strategy, we cannot disclose BERTaú, NER and Sentiment Analysis training datasets, as well as the complete FAQ dataset. However, we have assembled a small dataset of public FAQs from Itaú Unibanco bank website and we tackle this dataset with \texttt{BERTaú} pairwise model. The code with the FAQ experiment will be available soon at \url{https://github.com/itau/bertau}. 

%and can be tested in the public FAQs sample of data mentioned, or in any others datasets related to the tasks described in this paper.

%\bigskip
%\bigskip

\section{Drawbacks}
Here we briefly describe unsuccessful experiments

\begin{enumerate}[noitemsep]
    \item Train \texttt{BERTaú} from scratch starting from the final checkpoint of the \texttt{mBERT}: After $1,000,000$ steps the model performed very poorly when compared with the training from scratch without any initial checkpoint.
    \item FAQ retrieval experiment $-$ combine the \texttt{distiluse} and use \texttt{Faiss} \cite{faiss}: Did not go well when compared to the cosine distance solution.
    \item FAQ retrieval experiment $-$ use the \texttt{BM25+} to build the dataset and then use \texttt{BERTaú} as a re-ranking: There was no performance improvement and the solution only increased in complexity.
    \item FAQ retrieval experiment $-$ use \texttt{BERTaú}'s predictions for \texttt{BM25+} and Faiss: results were comparable with \texttt{BM25+} and worse with Faiss, our hypothesis is that the phrases ranked by \texttt{BERTaú} for a given query are more similar than a random choice, which may present greater difficulty for Faiss.
    \item NER experiment - A popular way to NER is using a \texttt{CRF} layer at the end of \texttt{BERT}, following the idea of \cite{bertimbau}. Our results with this idea weren't as good as the previously mentioned experiments, resulting in a slightly lower $F_1$ score and more symptoms of overfitting. Our hypothesis is that this result was influenced by our relatively small NER dataset.
\end{enumerate}

\end{document}